\documentclass{article}
\usepackage{spconf,amsmath,graphicx}

\usepackage{amsmath,amsfonts,amssymb,amsthm,dsfont,mathrsfs}
\usepackage{cite}
\usepackage{hyperref}
\usepackage{float}
\usepackage{amsfonts} 
\usepackage{bbm}

\newtheorem{remark}{Remark}

\def \bX {{\mathbf{X}}}
\def \bx {{\mathbf{x}}}

\def \bz {{\mathbf{z}}}

\def \bC {{\mathbf{C}}}

\def \bH {{\mathbf{H}}}

\def \bu {{\mathbf{u}}}

\def \cH {{\mathcal{H}}}

\usepackage{enumitem}
\setlist{nosep, leftmargin=14pt}

\usepackage{mwe} 

\title{Explainable Brain Age Gap Prediction in Neurodegenerative Conditions using coVariance Neural Networks}
 \name{Saurabh Sihag$^{\dagger}$  \qquad Gonzalo Mateos$^{\star}$ \qquad Alejandro Ribeiro$^{\ddagger}$}
\address{$^{\dagger}$ University at Albany, SUNY \\
$^{\star}$ University of Rochester \\
     $^{\ddagger}$University of Pennsylvania}
\begin{document}
%
\maketitle

\begin{abstract}
    Brain age is the estimate of biological age derived from neuroimaging datasets using machine learning algorithms. Increasing \textit{brain age gap} characterized by an elevated brain age relative to the chronological age can reflect increased vulnerability to neurodegeneration and cognitive decline. Hence, brain age gap is a promising biomarker for monitoring brain health. However, black-box machine learning approaches to brain age gap prediction have limited practical utility. Recent studies on coVariance neural networks (VNN) have proposed a relatively transparent deep learning pipeline for neuroimaging data analyses, which possesses two key features: (i) inherent \textit{anatomically interpretablity} of derived biomarkers; and (ii) a methodologically interpretable perspective based on \textit{linkage with eigenvectors of anatomic covariance matrix}.  In this paper, we apply the VNN-based approach to study brain age gap using cortical thickness features for various prevalent neurodegenerative conditions. Our results reveal distinct anatomic patterns for brain age gap in Alzheimer's disease, frontotemporal dementia, and atypical Parkinsonian disorders. Furthermore, we demonstrate that the distinct anatomic patterns of brain age gap are linked with the differences in how VNN leverages the eigenspectrum of the anatomic covariance matrix, thus lending explainability to the reported results. 
\end{abstract}

\section{Introduction}
Aging is a complicated biological process that manifests itself in the form of various progressive physiological and cognitive changes~\cite{lopez2013hallmarks}. Recent years have seen an exponential increase in the study of brain aging using machine learning algorithms~\cite{baecker2021machine}. A common objective of brain age prediction strategies is to derive an estimate of brain age from neuroimaging data and compare it with chronological age (time since birth) via the brain age gap, i.e., the difference between brain age and chronological age. A large age gap in the brain has been shown to be indicative of accelerated aging in various neurological diseases, implying a greater burden of the disease and the risk of mortality~\cite{jove2014metabolomics}. Subsequently, we use the notation $\Delta$-Age to refer to brain age gap.

The methodologies for inferring $\Delta$-Age in prior works follow the common template of (i) training a regression model to predict chronological age for a healthy population; and then (ii) application to a cohort of interest (often characterized by neurodegenerative condition), under the hypothesis that the trained model can detect `accelerated aging'~\cite{cole2017predicting, franke2019ten}. Existing studies in this domain have focused exclusively on (i) and there exist numerous machine learning models that have achieved impressive accuracy in predicting chronological age across heterogeneous populations of healthy individuals~\cite{bashyam2020mri, peng2021accurate, niu2020improved}. However, it is noteworthy that the residuals of the regression models that inform the $\Delta$-Age estimates are of \underline{primary interest} in this application, as the expectation is that they will drift in a specific direction when deployed to predict chronological age for individuals with adverse health conditions. Empirical evidence from several existing studies hints at decoupling the assessment of $\Delta$-Age from the prediction accuracy of chronological age on healthy population, as better performance in this task may not necessarily lead to a more informative $\Delta$-Age prediction~\cite{bashyam2020mri, jirsaraie2023systematic, sihag2023explainable}.  In this context, we build upon the recently proposed \textit{explanation-driven} pipeline for $\Delta$-Age prediction~\cite{sihag2023explainable} based on VNN models~\cite{sihag2022covariance} to provide an inherently transparent and \textit{anatomically interpretable} perspective to evaluation of $\Delta$-Age.

\subsection{coVariance Neural Networks}
VNNs have recently been studied as graph neural networks (GNNs) operating on the sample covariance matrix as the graph~\cite{sihag2022covariance,cavallo2024fair, cavallo2024sparsecovarianceneuralnetworks, cavallo2024spatiotemporal, sihagJSTSP, sihag2023explainable}. VNNs belong to the convolutional family of deep learning models as they rely on \emph{linear-shift-and-sum} operators~\cite{ortega2018graph} on the covariance matrix. Specifically, given a sample covariance matrix ${\bC} \in \mathbb{R}^{m\times m}$, the convolution operation in a VNN is modeled by a \underline{coVariance filter}, given by $\bH(\bC) \triangleq \sum_{k=0}^K h_k \bC^k$, where scalar parameters $\{h_k\}_{k=0}^K$ are referred to as filter taps that are learned from the data. The application of coVariance filter $\bH(\bC)$ on an input $\bx$ translates to combining information across different sized neighborhoods. \textit{VNNs have been studied separately from GNNs} because they draw a fundamental equivalence with principal component analysis (PCA) when $\bC$ is the sample covariance matrix~\cite{sihag2022covariance, cavallo2024sparsecovarianceneuralnetworks, cavallo2024spatiotemporal}. Advantages of VNNs over traditional PCA-based approaches include improved performance, stability~\cite{sihag2022covariance, cavallo2024spatiotemporal}, and, {notably}, transferability across multiscale datasets~\cite{sihagJSTSP}. In this context, VNNs are deep learning models that \textit{bridge the gap} between traditional PCA-based statistical approaches and graph neural networks~\cite{sihag2022covariance, sihag2023explainable, sihagJSTSP, cavallo2024fair, cavallo2024spatiotemporal}.

\subsection{VNNs for neuroimaging data analysis}
VNNs have been successfully used to extract biomarkers of neurodegeneration in Alzheimer's disease from brain morphometric features~\cite{sihag2023explainable, sihagJSTSP, sihag2024towards, sihag2022predicting}. VNNs offer two key features in the context of deriving biomarkers from neuroimaging data: (i) Representations learned by VNNs provide anatomic signatures of biomarkers~\cite{sihag2023explainable}; and (ii) learning outcomes of VNNs are generalizable to neuroimaging data  curated according to other brain atlases under certain regularity conditions~\cite{sihagJSTSP}. In particular, VNN-based data analysis pipeline has yielded a novel \textit{explanation-driven} perspective to brain age gap gap evaluation~\cite{sihag2023explainable}, which is distinct from the prevalent machine learning works in this domain that leverage black-box models and focus primarily on the performance of chronological age prediction in healthy individuals for gauging their quality.  

\subsection{Contributions}
In this paper, we leverage the brain age gap prediction pipeline based on VNNs to study $\Delta$-Age from cortical thickness features and its explainability for Alzheimer's disease (AD), frontotemporal degeneration (FTD), as well as corticobasal syndrome and progressive supranuclear palsy (together referred as the cohort of atypical Parkinsonian disorders (APD) in this paper), and Parkinsons' disease (PD); the data for all cohorts were downloaded from~\href{https://ida.loni.usc.edu}{https://ida.loni.usc.edu}. Our contributions are summarized as follows:
\begin{itemize}
    \item[-] {\bf Anatomical characterization of $\Delta$-age for different neurodegenerative conditions.} Our results demonstrated significantly elevated $\Delta$-Age for AD, FTD, and APD conditions relative to a healthy population, along with distinct and biologically plausible anatomic patterns associated with $\Delta$-Age for these conditions. 
    \item[-] {\bf Explainability of $\Delta$-Age.} Our experiments demonstrated that a pre-trained VNN model exploited the eigenvectors of the anatomical covariance matrix differently for the aforementioned neurodegenerative conditions, thus rendering explainability to the distinct anatomic patterns associated with $\Delta$-Age. 
\end{itemize}
Previous work on VNN-based $\Delta$-Age prediction has focused primarily on Alzheimer's disease~\cite{sihag2022predicting, sihag2023explainable, sihagJSTSP}. This paper provides a more comprehensive evaluation than prior works, as it demonstrates the applicability of VNNs for the prediction of $\Delta$-Age for several neurodegenerative conditions and elucidates the interpretability and explainability of $\Delta$-Age across them. 

\section{coVariance Neural Networks}
\begin{figure}[t]
    \centering
    \includegraphics[width=0.75\linewidth]{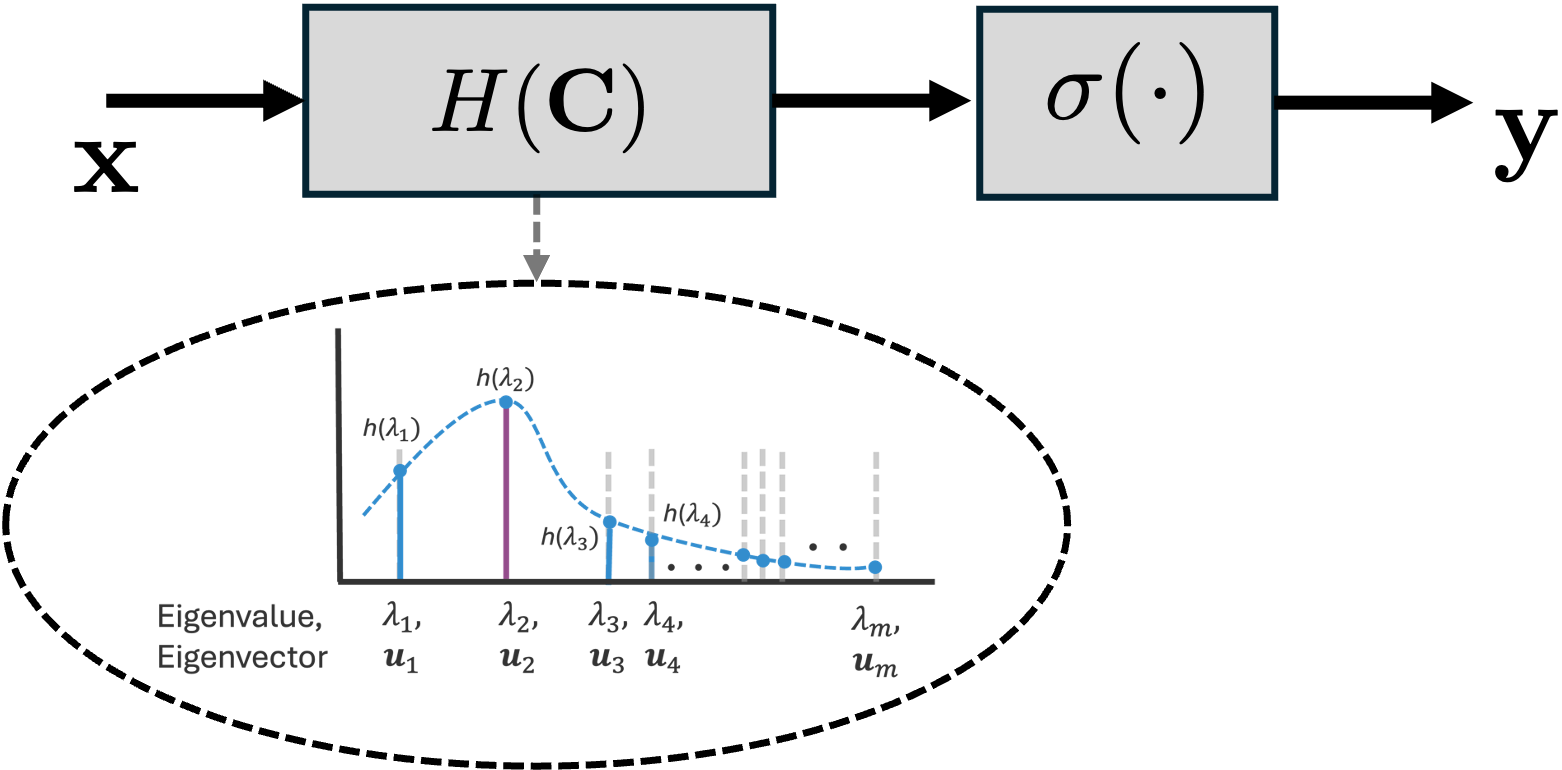}
    \caption{One layer of a VNN model. The covariance filter implicitly manipulates input data according to the eigenvectors of the covariance matrix $\bC$, thus tying the output ${\bf y}$ to specific eigenvectors of $\bC$.}
    \label{fig:vnn1layer}
\end{figure}

We start by providing a brief overview of the architecture of a VNN model. A single layer of VNN is formed by concatenating a coVariance filter with a pointwise non-linear activation function $\sigma(\cdot)$ (e.g., ${\sf ReLU}, \tanh$) that satisfies $\sigma(\bu) = [\sigma(u_1), \dots, \sigma(u_m)]$ for $\bu = [u_1, \dots, u_m]$. Therefore, the output of a single layer VNN with input $\bx$ is given by $\bz = \sigma(\bH(\bC) \bx)$ (Fig.~\ref{fig:vnn1layer}). The construction of a multi-layer VNN is formalized next.

\begin{remark}[Multi-layer VNN]
For an $L$-layer VNN, denote the coVariance filter in layer $\ell$ of the VNN by~$\bH_{\ell}(\bC)$ and its corresponding set of filter taps by $\cH_{\ell}$. Given a pointwise nonlinear activation function $\sigma(\cdot)$, the relationship between the input $\bx_{\ell-1}$ and the output $\bx_{\ell}$ for the $\ell$-th layer is $\bx_{\ell} = \sigma(\bH_{\ell}(\bC)\bx_{\ell-1})\,\quad\text{for}\quad \ell\in \{1,\dots,L\}$, where $\bx_0$ is the input $\bx$. 
 \end{remark}
Furthermore, similar to other deep learning models, sufficient expressive power can be facilitated in the VNN architecture by incorporating multiple input multiple output (MIMO) processing at every layer. Formally, consider a VNN layer $\ell$ that can process $F_{\ell-1}$ number of $m$-dimensional inputs and outputs $F_{\ell}$ number of $m$-dimensional outputs via $F_{\ell-1} \times F_{\ell}$ number of filter banks~\cite{gama2020stability}. In this scenario, the input is specified as $\bX_{\sf in} = [\bx_{\sf in}[1],\dots,\bx_{\sf in}[F_{\sf in}]]$, and the output is specified as $\bX_{\sf out} = [\bx_{\sf out}[1],\dots,\bx_{\sf out}[F_{\sf out}]]$. The relationship between the $f$-th output $\bx_{\sf out}[f]$ and the input $\bx_{\sf in}$ is given by $\bx_{\sf out}[f] =  \sigma\Big(\sum_{g = 1}^{F_{\sf in} } \bH_{fg}(\bC)\bx_{\sf in} [g] \Big)$,
where $\bH_{fg}(\bC)$ is the coVariance filter that processes $\bx_{\sf in}[g]$. Without loss of generality, we assume that $F_{\ell} = F,\forall \ell \in \{1,\dots,L\}$. In this case, the set of all filter taps is given by  ${\cal H} = \{\cH_{fg}^{\ell}\}, \forall f,g \in \{1,\dots, F\}, \ell \in \{1,\dots,L\}$, where $\cH_{fg} = \{h_{fg}^{\ell}[k]\}_{k=0}^K$ and $h_{fg}^{\ell}[k]$ is the $k$-th filter tap for filter $\bH_{fg}(\bC)$. Thus, we can compactly represent a multi-layer VNN architecture capable of MIMO processing via the notation $\Phi(\bx;\bC,{\cal H})$, where the set of filter taps $\cH$ captures the full span of its architecture. 
\begin{remark}[Statistical inference using VNNs]\label{vnnpca}
    Theorem 1 in~\cite{sihag2022covariance} established the equivalence between processing data samples with principal component analysis (PCA) transform and processing data samples with a coVariance filter $\bH(\bC)$. Hence, it can be concluded that input data is processed with VNNs, at least in part, by exploiting the eigenvectors of~$\bC$. Unlike simpler PCA-based inference models, VNNs offer stability~\cite{sihag2022covariance} and transferability guarantees~\cite{sihag2022predicting}, which ensure reproducibility of the inference outcomes by VNNs with high confidence.
\end{remark}
\section{Methods and Materials}

\subsection{Datasets for neurodegenerative conditions}\label{data}
In this paper, we leverage cortical thickness measures derived from structural MRI and curated according to Desikan-Killiany brain atlas ($68$ cortical regions) for various neurodegenerative conditions. All datasets are publicly available at \href{https://ida.loni.usc.edu/}{https://ida.loni.usc.edu/}.

\noindent
{\bf ADNI.}  This dataset comprised of 118 individuals diagnosed with Alzheimer's disease dementia (AD; age = $73.84\pm 7.56$ years, $56$ females) and $206$ healthy individuals (HC; age = $73.87\pm 6.39$ years, $110$ females). The $68$ cortical thickness features across the cortex for all individuals were downloaded from
\href{https://adni.loni.usc.edu/}{https://adni.loni.usc.edu/}. The cortical thickness features had been derived from T1w MRI images using Freesurfer 5.1. This dataset was collected as part of the Alzheimer's disease neuroimaging iniative (ADNI)~\cite{jack2008alzheimer}. 

\noindent 
{\bf NIFD.} This dataset spans $119$ individuals diagnosed with different forms of FTD (FTD; age = $64.72\pm 6.78$ years, $47$ females). In the FTD group, $52$ individuals were diagnosed with behavioral variant FTD (BV; age = $63.07\pm 5.77$ years, $16$ females); $36$ individuals were diagnosed with semantic variant of primary progressive aphasia (SV; age =  $63.68\pm 6.17$ years, $15$ females); and $31$ individuals with non-fluent variant of primary progressive aphasia (PNFA; age =  $68.7\pm 7.61$ years, $16$ females). Additionally, this dataset consisted of $114$ healthy individuals (HC; age = $64.51\pm 6.5$ years, $65$ females). The MRI images were collected as part of the frontotemporal lobar degeneration neuroimaging initiative (FTLDNI). Cortical thickness features (curated according to Desikan-Killiany atlas) were derived using the open-access CAT12 pipeline~\cite{gaser2022cat} using their default options. We refer the reader to \href{https://neuro-jena.github.io/cat12-help/}{https://neuro-jena.github.io/cat12-help/} for detailed processing steps. All outputs
were quality checked visually for errors in grey matter segmentation. 

\noindent 
{\bf 4RTNI.} This dataset was collected as part of the 4-Repeat Tauopathy Neuroimaging Initiative (4RTNI) and used similar MRI acquisition and clinical assessments as the NIFD dataset. This dataset constituted of 59 individuals diagnosed with progressive supranuclear palsy (PSP; age = $70.79 \pm 7.65$ years, $32$ females) and $45$ individuals diagnosed with corticobasal syndrome (CBS; age = $66.71\pm 6.75$ years, $24$ females). CBS and PSP disorders are among the most common disorders within the broader family of APD~\cite{fogel2014neurogenetics}. In this paper, we denote the combined cohort of CBS and PSP as APD group. Cortical thickness features curated according to the Desikan-Killiany atlas were derived from T1w MRI using similar pre-processing steps as that for the NIFD dataset. Furthermore, the HC group from the NIFD dataset was considered as the healthy control group for 4RTNI dataset due to the similarity in the MRI acquisition methods of NIFD and 4RTNI  datasets.

\noindent
{\bf PPMI.} This dataset constituted of $454$ individuals diagnosed with Parkinson's disease (PD; age = $63.17\pm 9.37$ years, $174$ females) and $133$ healthy individuals (HC; age = $61.44\pm 11.45$ years, $45$ females). Cortical thickness features curated according to the Desikan-Killiany atlas were derived from T1w MRI using similar pre-processing steps as that for the NIFD dataset. 
\subsection{$\Delta$-Age prediction}
\noindent
{\bf VNN architecture and training.} The VNN model was pre-trained on a healthy population to glean information about healthy aging. To facilitate $\Delta$-Age that is transparent and methodologically interpretable, we used a multi-layer VNN model that yielded representations from the input cortical thickness features at the final layer, such that the unweighted mean of these representations formed the estimate for chronological age. Details on how this choice of architecture leads to anatomically interpretable $\Delta$-Age are discussed subsequently.
\begin{figure}[t]
    \centering
    \includegraphics[width=0.9\linewidth]{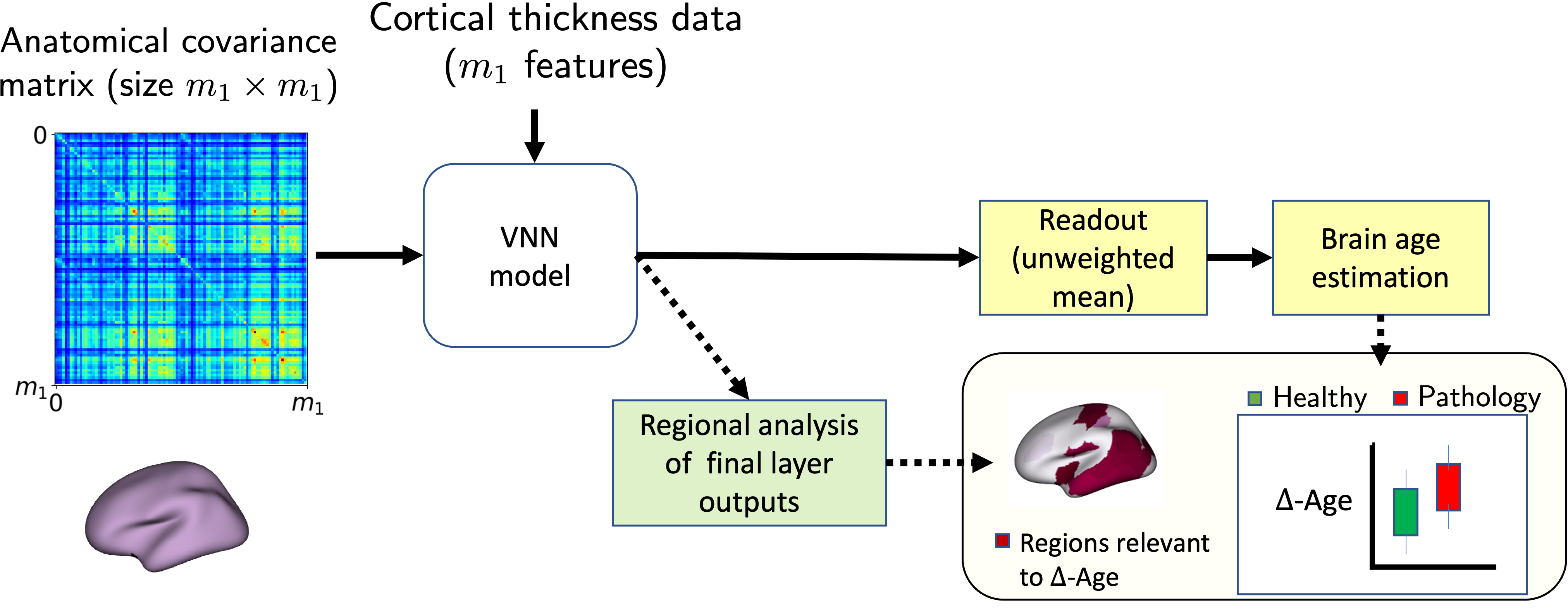}
    \caption{A VNN-based pipeline for $\Delta$-Age prediction.}
    \label{fig:vnnage}
\end{figure}
The VNN model consisted of two layers and yielded a representation from the input cortical thickness data via transformation that was dictated by the anatomical covariance matrix.  For training this VNN model, we leveraged the cortical thickness features from the healthy control population in the publicly available OASIS-3 dataset~\cite{lamontagne2019oasis}. The healthy population in the OASIS-3 dataset consisted of $631$ individuals (age = $67.71\pm 8.37$ years, $367$ females). 

For training, we randomly split the available cortical thickness data into a training set of $568$ individuals and a test set of $63$ individuals. The anatomical covariance matrix was estimated from the cortical thickness data of the training set. The training set was further split into a subset of $498$ individuals and a validation set of $70$ individuals. The VNN was trained to predict chronological age on the subset of $498$ individuals with mean squared error loss optimized using stochastic gradient descent with Adam optimizer for up to $100$ epochs. The configuration with the best performance on the validation set of $70$ individuals was selected.

The first layer of VNN consisted of $2$ filter taps and the second layer consisted of $6$ filter taps, with width $61$. Thus, in total, VNN model consisted of $22,570$ learnable parameters. The batch size used for training was $10$ and the learning rate was $0.15$. The hyperparameters for the VNN architecture and training were decided during a hyperoptimization procedure based on Optuna~\cite{akiba2019optuna}.  Using this strategy, we trained $10$ distinct VNN models with different permutations of the training set. These models achieved a prediction performance of $7.25\pm 0.51$ years on the test set and $6.33$ years on the complete dataset, with a Pearson's correlation of $0.44\pm0.014$. Thus, the statistical evidence suggested that VNNs learned information about healthy aging, even though they were weak predictors of chronological age. The results reported in this paper are derived from one pre-trained VNN model among the $10$ that were pre-trained using the above procedure.

\noindent
{\bf Anatomically interpretable and explainable $\Delta$-Age prediction.} The pre-trained VNN model facilitated the prediction of $\Delta$-Age and associated anatomic interpretability and explainability for the cohorts associated with neurodegenerative conditions in Section~\ref{data}. Figure~\ref{fig:vnnage} provides an overview of VNN-based pipeline for $\Delta$-Age prediction.
\begin{itemize}
    \item \textit{Evaluating $\Delta$-Age.} $\Delta$-Age is evaluated as the difference between brain age and chronological age. Brain age was evaluated from the chronological age estimates formed by the VNN with application of a standard linear regression-based approach~\cite{beheshti2019bias}; the weights of linear regression model learned from the HC populations in the respective datasets in Section~\ref{data}. 
    \item \textit{Assigning anatomical interpretability to $\Delta$-Age.} Elevated $\Delta$-Age in disease groups could be attributed to the statistical patterns in the representations formed by the pre-trained VNN in its final layer. Since the convolution operations in VNN preserve the original dimensionality of the input data, the representations at the final layer could be mapped to individual brain regions. Hence, we evaluated a set of \underline{regional residuals} $r_i$ for each brain region $i$, which were defined as the difference between the chronological estimate and the output in the final layer of the VNN corresponding to that brain region. The elevations in the regional residuals directly contribute to elevated $\Delta$-Age~\cite[Section 3.3]{sihag2023explainable}. 
    \item \textit{Explainability of $\Delta$-Age.} The representations are learned by the VNN, \textit{in part},  by transforming the input data according to the eigenspectrum of the anatomical covariance matrix~\cite{sihag2022covariance}. By leveraging this fact, we characterize the explainability of $\Delta$-Age by evaluating the \underline{inner products} between the regional residuals derived from representations learned by the VNN and the eigenvectors of the anatomical covariance matrix. We anticipate to observe significant differences in terms of these inner product metrics for disease groups and healthy populations. These experiments will elucidate how VNN processed the cortical thickness information from disease groups differently relative to the healthy population, thus lending \textit{explainability} to the evaluation of the downstream statistic of $\Delta$-Age in different cohorts. 
\end{itemize}

\begin{figure}[t]
  \centering
  \includegraphics[scale=0.15]{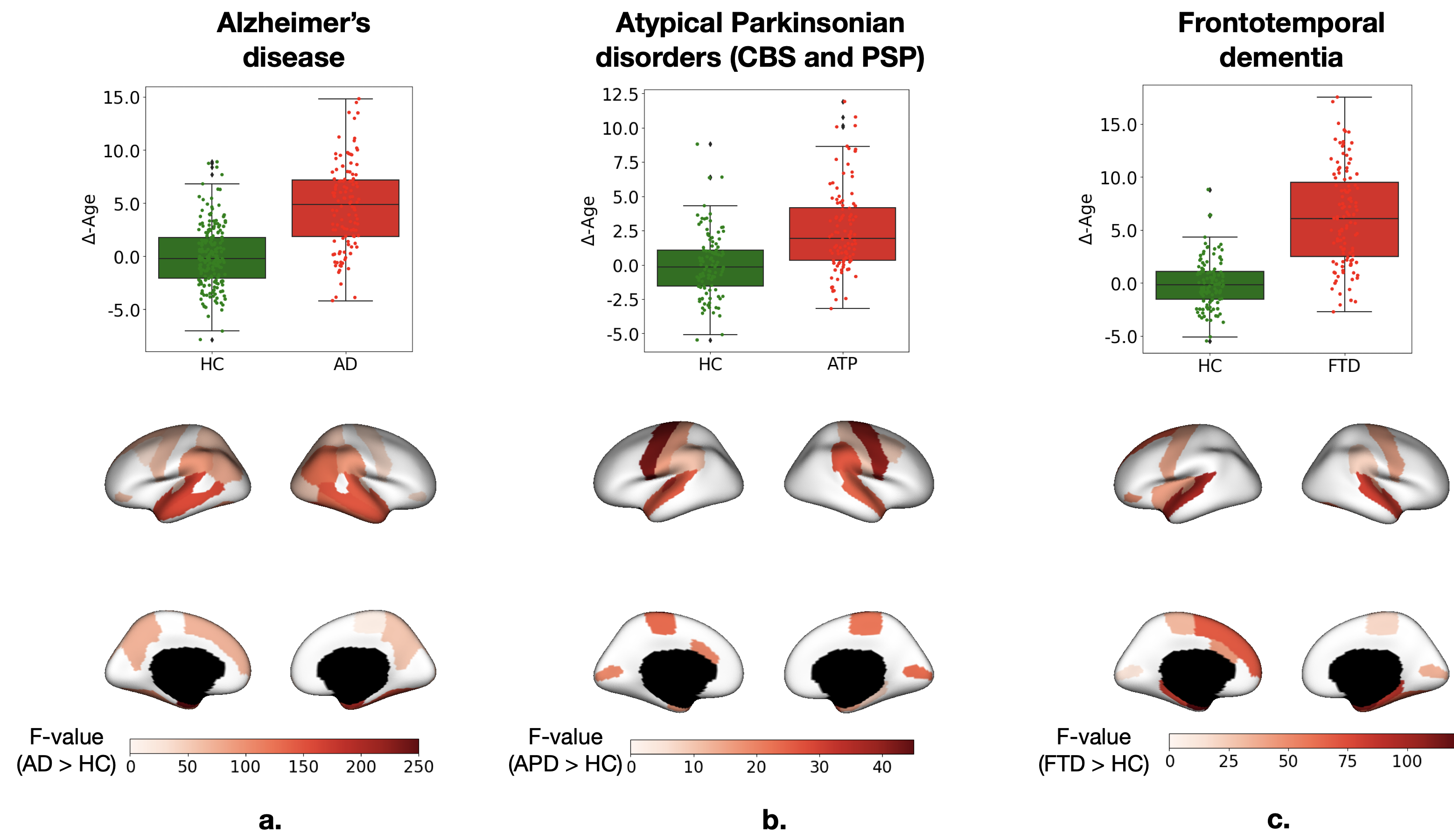}
   \caption{Brain age estimates and associated anatomic characterizations for ({\bf a}) AD, ({\bf b}) APD, and ({\bf c}) FTD.}
   \label{neurovnn_age}
\end{figure}

\section{Results}
For each disease dataset, we used the anatomical covariance matrix estimated only from the respective HC group in the pre-trained VNN model. VNN was oblivious to the identity or any information about the disease. Figure~\ref{neurovnn_age} illustrates that elevated $\Delta$-Age was observed for AD, ATP, and FTD relative to their respective HC groups. The {\bf $\Delta$-Age for AD group} was $4.67\pm 4.04$ years, which was elevated relative to the HC group in this dataset ($\Delta$-Age for HC: $0\pm 2.91$ years). Figure~\ref{neurovnn_age}a also illustrates the anatomic characterization of elevated $\Delta$-Age in AD group. The anatomic characterization was derived by plotting the $F$-values for ANOVA between individual elements of the representations generated by the VNN model for AD and HC groups on the brain surface (only for group differences that had AD $>$ HC and survived Bonferroni correction for multiple comparisons with $p$-value $<0.05$). The directionality AD $>$ HC is relevant to $\Delta$-Age as the rise in the individual elements of the representations learned by VNN for a disease group contributed to the elevated $\Delta$-Age estimate relative to healthy population~\cite{sihag2023explainable}. The anatomic characterization for $\Delta$-Age in AD spanned bilateral regions in the medial temporal lobe, entorhinal, and temporo-parietal junction, which are relevant regions for AD pathology~\cite{desikan2009temporoparietal}. Consistent results were observed in the prior work on VNN-based $\Delta$-Age prediction~\cite{sihag2023explainable}.

\begin{figure}[t]
  \centering
  \includegraphics[scale=0.15]{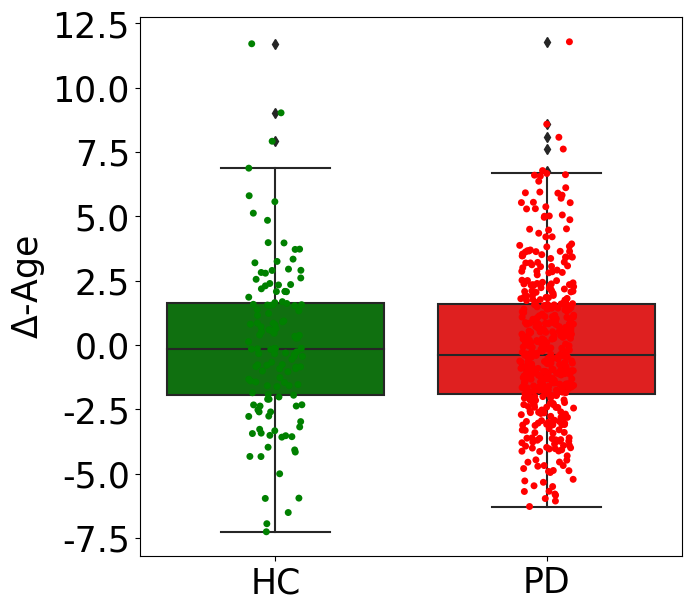}
   \caption{$\Delta$-Age for PD and respective HC group.}
   \label{pd_age}
\end{figure}

Similar analysis procedures were followed for 4RTNI, NIFD, and PPMI datasets, and the associated results for APD and FTD disease groups are illustrated in Fig.~\ref{neurovnn_age}b and Fig.~\ref{neurovnn_age}c, respectively. The HC group for 4RTNI and NIFD datasets was the same and had $\Delta$-Age $= 0\pm 2.33$ years. The $\Delta$-Age for FTD disease group was $6.17\pm 4.55$ years and that for the APD group was $2.49\pm 3.09$ years, suggesting that FTD group exhibited more significant accelerated aging. The {\bf $\Delta$-Age in FTD group} was characterized by superior frontal region in the left hemisphere, bilateral entorhinal, parahippocampal, and superior temporal regions, and relatively less prominent contributions from bilateral precentral regions, which are part of the motor cortex. FTD is characterized by neurodegeneration in the regions in frontal and temporal lobes~\cite{rohrer2012structural}, and hence, the anatomic characterization in Fig.~\ref{neurovnn_age}c captures their impact in the form of elevated $\Delta$-Age. The {\bf $\Delta$-Age in APD group} was characterized by brain regions comprising bilateral superior temporal, precentral, and occipital lobes. The CBS and PSP pathologies in APD group exhibit symptoms similar to PD in terms of motor dysfunction, but are also characterized by cognitive dysfunction and typically rapid decline in function relative to PD~\cite{mcfarland2016diagnostic}. Hence, the implication of regions in the motor cortex and occipital lobe in Fig.~\ref{neurovnn_age}b is relevant to the disease characteristic in APD. Interestingly, unlike the other disease groups, no significant difference in $\Delta$-Age was observed in the PD group relative to its respective HC group ({Fig.~\ref{pd_age}}). 

\begin{figure}[t]
    \centering
    \includegraphics[width=\linewidth]{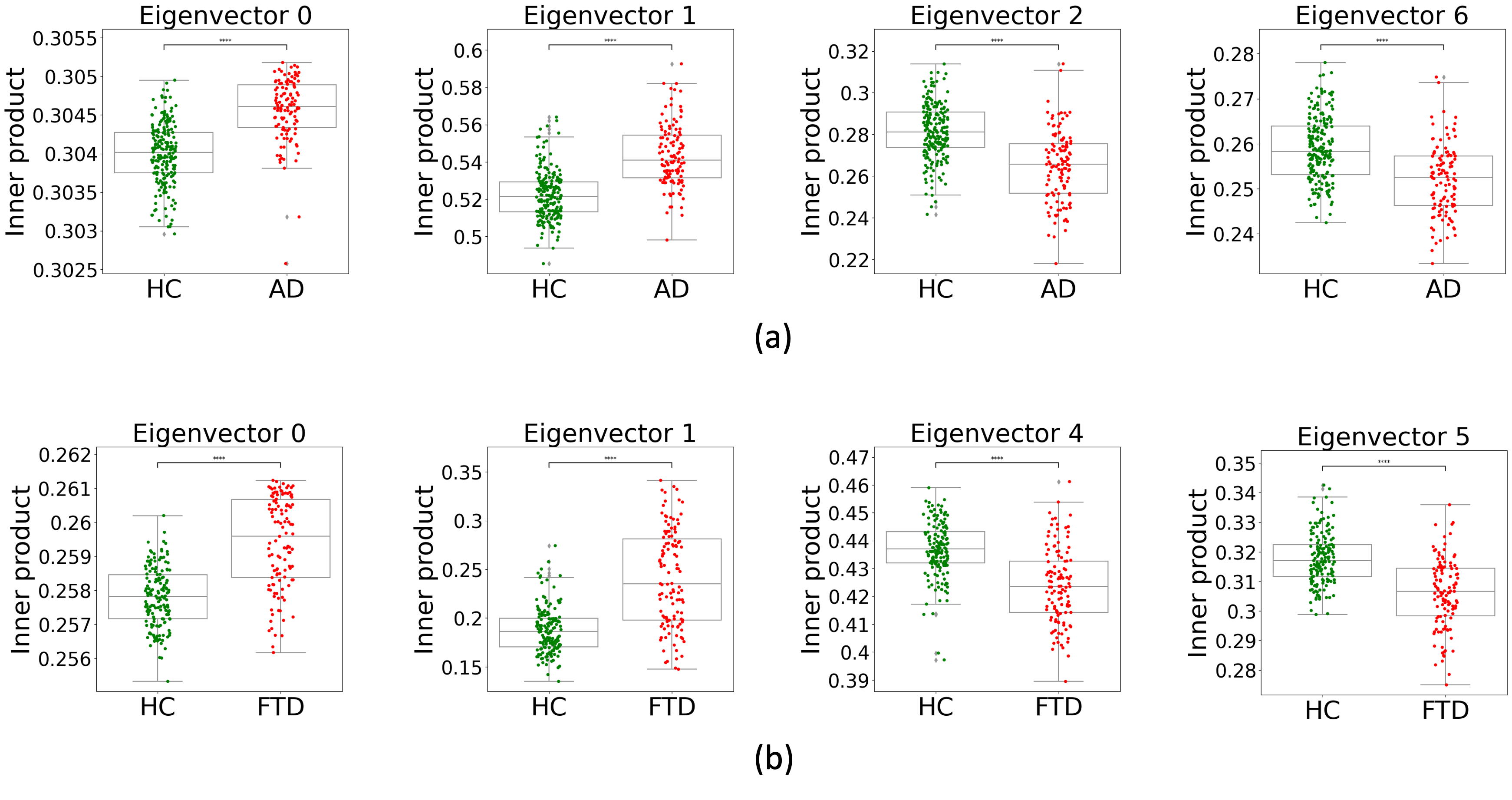}
    \caption{Explaining $\Delta$-Age in disease groups (({\bf a}) AD and ({\bf b}) FTD) in terms of the group differences between inner products of VNN representations and eigenvectors of anatomical covariance matrices. ($^{\ast\ast\ast\ast}:$ $p$-value $\leq 1\exp-4$) }
    \label{fig:expl}
\end{figure}

\noindent
{\bf Explainability of $\Delta$-Age.} Next, we analyzed the inner products between the regional residuals derived from the representations learned by VNNs and the eigenvectors of the anatomical covariance matrix. The eigenvectors of the anatomical covariance matrix were organized from $0$ to $67$, with the eigenvector $0$ associated with the largest eigenvalue and the $67$-th eigenvector associated with the smallest eigenvalue. The inner product metrics were significantly different (ANOVA, $p$-value $<0.0001$) between the AD group and HC groups for the eigenvectors $0$, $1$, $2$, and $6$ of the anatomical covariance matrix (Fig.~\ref{fig:expl}a). Thus, the VNN model processed the cortical thickness features for AD group significantly different relative to the HC group leading to distinct distributions in $\Delta$-Age in Fig.~\ref{neurovnn_age}a, and these differences were dictated by variations in how the VNN exploited the eigenvectors of the anatomical covariance matrix for AD and HC groups. 

Similar results were obtained for the FTD cohort, where the most significant group differences in the inner product measures were observed for eigenvectors $0, 1,4$, and $5$ (Fig.~\ref{fig:expl}b). Thus, the variations in how the VNN exploited the eigenvectors of the anatomical covariance matrix for FTD and HC groups \textit{explained} the variations in $\Delta$-Age in Fig.~\ref{neurovnn_age}c. Notably, the inner product measures were significantly different between the APD and HC groups only for eigenvector $8$, which explains the relatively smaller elevation in $\Delta$-Age in the APD group relative to FTD or AD groups in Fig.~\ref{neurovnn_age}.

\section{Discussion}
The results in Fig.~\ref{neurovnn_age} illustrate that the distributions of $\Delta$-Age estimates can be substantially overlapping across different diseases. Hence, $\Delta$-Age, by itself, is not a sufficient indicator to characterize a disease. In this context, the anatomic characterization of $\Delta$-Age offered by VNN embellishes its informative aspect about neurodegeneration. Notably, the anatomic characterizations of $\Delta$-Age for AD, FTD, and ATP disease groups were unique and characteristic of the respective diseases.

\noindent
{\bf Comparison with existing literature.} Existing studies in this domain are focused primarily on brain age prediction, which is to be contrasted with this paper's focus on $\Delta$-Age prediction. Moreover, existing studies utilize the state-of-the-art post-hoc, model-agnostic methods, such as, SHAP, LIME~\cite{lombardi2021explainable}, saliency maps~\cite{yin2023anatomically}, and layer-wise relevance propagation~\cite{hofmann2022towards} to \textit{explain} the brain age predictions. These methods add anatomical interpretability to brain age estimates by assigning some importance to the input features (often associated with specific anatomic regions). Unlike these approaches, we leverage the \textit{inherent explainability} of the VNN model and our results bring into focus the \textit{the properties that a VNN gains when it is exposed to the information provided by chronological age of healthy controls} and \textit{whether and how these properties translate to a meaningful $\Delta$-Age estimate}. 

\noindent
{\bf Future work.} Future work could entail clinical validation of reported $\Delta$-Age measures by exploring their associations with various clinical markers of a disease. Moreover, longitudinal analyses could provide insights into how $\Delta$-Age progresses with disease burden.

\section{Acknowledgement}
Data used in the preparation of this manuscript were obtained from the Alzheimer's Disease Neuroimaging Initiative (ADNI) database, the Frontotemporal Lobar Degeneration Neuroimaging Initiative (FTLDNI) database, and Parkinson’s
Progression Markers Initiative (PPMI) database.

{\small 
\bibliographystyle{IEEEbib}
\bibliography{VNN_ICML}

\begin{thebibliography}{10}

\bibitem{lopez2013hallmarks}
Carlos L{\'o}pez-Ot{\'\i}n et~al.,
\newblock ``The hallmarks of aging,''
\newblock {\em Cell}, vol. 153, no. 6, pp. 1194--1217, 2013.

\bibitem{baecker2021machine}
Lea Baecker, Rafael Garcia-Dias, Sandra Vieira, Cristina Scarpazza, and Andrea
  Mechelli,
\newblock ``Machine learning for brain age prediction: Introduction to methods
  and clinical applications,''
\newblock {\em EBioMedicine}, vol. 72, pp. 103600, 2021.

\bibitem{jove2014metabolomics}
Mariona Jov{\'e}, Manuel Portero-Ot{\'\i}n, Alba Naud{\'\i}, Isidre Ferrer, and
  Reinald Pamplona,
\newblock ``Metabolomics of human brain aging and age-related neurodegenerative
  diseases,''
\newblock {\em Journal of Neuropathology \& Experimental Neurology}, vol. 73,
  no. 7, pp. 640--657, 2014.

\bibitem{cole2017predicting}
James~H Cole and Katja Franke,
\newblock ``Predicting age using neuroimaging: Innovative brain ageing
  biomarkers,''
\newblock {\em Trends in Neurosciences}, vol. 40, no. 12, pp. 681--690, 2017.

\bibitem{franke2019ten}
Katja Franke and Christian Gaser,
\newblock ``Ten years of brainage as a neuroimaging biomarker of brain aging:
  What insights have we gained?,''
\newblock {\em Frontiers in neurology}, p. 789, 2019.

\bibitem{bashyam2020mri}
Vishnu~M Bashyam, Guray Erus, Jimit Doshi, Mohamad Habes, Ilya~M Nasrallah,
  Monica Truelove-Hill, Dhivya Srinivasan, Liz Mamourian, Raymond Pomponio,
  Yong Fan, et~al.,
\newblock ``{MRI} signatures of brain age and disease over the lifespan based
  on a deep brain network and 14 468 individuals worldwide,''
\newblock {\em Brain}, vol. 143, no. 7, pp. 2312--2324, 2020.

\bibitem{peng2021accurate}
Han Peng, Weikang Gong, Christian~F Beckmann, Andrea Vedaldi, and Stephen~M
  Smith,
\newblock ``Accurate brain age prediction with lightweight deep neural
  networks,''
\newblock {\em Medical image analysis}, vol. 68, pp. 101871, 2021.

\bibitem{niu2020improved}
Xin Niu, Fengqing Zhang, John Kounios, and Hualou Liang,
\newblock ``Improved prediction of brain age using multimodal neuroimaging
  data,''
\newblock {\em Human brain mapping}, vol. 41, no. 6, pp. 1626--1643, 2020.

\bibitem{jirsaraie2023systematic}
Robert~J Jirsaraie, Aaron~J Gorelik, Martins~M Gatavins, Denis~A Engemann, Ryan
  Bogdan, Deanna~M Barch, and Aristeidis Sotiras,
\newblock ``A systematic review of multimodal brain age studies: Uncovering a
  divergence between model accuracy and utility,''
\newblock {\em Patterns}, vol. 4, no. 4, 2023.

\bibitem{sihag2023explainable}
Saurabh Sihag, Gonzalo Mateos, Corey McMillan, and Alejandro Ribeiro,
\newblock ``Explainable brain age prediction using covariance neural
  networks,''
\newblock in {\em Thirty-seventh Conference on Neural Information Processing
  Systems}, 2023.

\bibitem{sihag2022covariance}
Saurabh Sihag, Gonzalo Mateos, Corey McMillan, and Alejandro Ribeiro,
\newblock ``{coVariance} neural networks,''
\newblock in {\em Proc. Conference on Neural Information Processing Systems},
  Nov. 2022.

\bibitem{cavallo2024fair}
Andrea Cavallo, Madeline Navarro, Santiago Segarra, and Elvin Isufi,
\newblock ``Fair covariance neural networks,''
\newblock {\em arXiv preprint arXiv:2409.08558}, 2024.

\bibitem{cavallo2024sparsecovarianceneuralnetworks}
Andrea Cavallo, Zhan Gao, and Elvin Isufi,
\newblock ``Sparse covariance neural networks,''
\newblock {\em arXiv:2410.01669}, vol. cs.LG, 2024.

\bibitem{cavallo2024spatiotemporal}
Andrea Cavallo, Mohammad Sabbaqi, and Elvin Isufi,
\newblock ``Spatiotemporal covariance neural networks,''
\newblock in {\em Joint European Conference on Machine Learning and Knowledge
  Discovery in Databases}. Springer, 2024, pp. 18--34.

\bibitem{sihagJSTSP}
Saurabh Sihag, Gonzalo Mateos, Corey McMillan, and Alejandro Ribeiro,
\newblock ``Transferability of covariance neural networks,''
\newblock {\em IEEE Journal of Selected Topics in Signal Processing}, pp.
  1--16, 2024.

\bibitem{ortega2018graph}
Antonio Ortega, Pascal Frossard, Jelena Kova{\v{c}}evi{\'c}, Jos{\'e}~MF Moura,
  and Pierre Vandergheynst,
\newblock ``Graph signal processing: Overview, challenges, and applications,''
\newblock {\em Proceedings of the IEEE}, vol. 106, no. 5, pp. 808--828, 2018.

\bibitem{sihag2024towards}
Saurabh Sihag, Gonzalo Mateos, and Alejandro Ribeiro,
\newblock ``Towards a foundation model for brain age prediction using
  covariance neural networks,''
\newblock {\em arXiv preprint arXiv:2402.07684}, 2024.

\bibitem{sihag2022predicting}
Saurabh Sihag, Gonzalo Mateos, Corey McMillan, and Alejandro Ribeiro,
\newblock ``Predicting brain age using transferable {coVariance} neural
  networks,''
\newblock in {\em Proc. IEEE International Conference on Acoustics, Speech, and
  Signal Processing}, Jun. 2023.

\bibitem{gama2020stability}
Fernando Gama, Joan Bruna, and Alejandro Ribeiro,
\newblock ``Stability properties of graph neural networks,''
\newblock {\em IEEE Transactions on Signal Processing}, vol. 68, pp.
  5680--5695, 2020.

\bibitem{jack2008alzheimer}
Clifford~R Jack~Jr, Matt~A Bernstein, Nick~C Fox, Paul Thompson, Gene
  Alexander, Danielle Harvey, Bret Borowski, Paula~J Britson, Jennifer
  L.~Whitwell, Chadwick Ward, et~al.,
\newblock ``The alzheimer's disease neuroimaging initiative (adni): Mri
  methods,''
\newblock {\em Journal of Magnetic Resonance Imaging: An Official Journal of
  the International Society for Magnetic Resonance in Medicine}, vol. 27, no.
  4, pp. 685--691, 2008.

\bibitem{gaser2022cat}
Christian Gaser, Robert Dahnke, Paul~M Thompson, Florian Kurth, Eileen Luders,
  and {A}lzheimer’s Disease Neuroimaging~Initiative,
\newblock ``{CAT}--a computational anatomy toolbox for the analysis of
  structural mri data,''
\newblock {\em biorxiv}, pp. 2022--06, 2022.

\bibitem{fogel2014neurogenetics}
Brent~L Fogel, Mary~C Clark, and Daniel~H Geschwind,
\newblock ``The neurogenetics of atypical parkinsonian disorders,''
\newblock in {\em Seminars in neurology}. Thieme Medical Publishers, 2014,
  vol.~34, pp. 217--224.

\bibitem{lamontagne2019oasis}
Pamela~J LaMontagne, Tammie~LS Benzinger, John~C Morris, Sarah Keefe, Russ
  Hornbeck, Chengjie Xiong, Elizabeth Grant, Jason Hassenstab, Krista Moulder,
  Andrei~G Vlassenko, et~al.,
\newblock ``{OASIS}-3: longitudinal neuroimaging, clinical, and cognitive
  dataset for normal aging and {A}lzheimer disease,''
\newblock {\em MedRxiv}, 2019.

\bibitem{akiba2019optuna}
Takuya Akiba, Shotaro Sano, Toshihiko Yanase, Takeru Ohta, and Masanori Koyama,
\newblock ``Optuna: A next-generation hyperparameter optimization framework,''
\newblock in {\em Proceedings of the 25th ACM SIGKDD International Conference
  on Knowledge Discovery \& Data Mining}, 2019, pp. 2623--2631.

\bibitem{beheshti2019bias}
Iman Beheshti, Scott Nugent, Olivier Potvin, and Simon Duchesne,
\newblock ``Bias-adjustment in neuroimaging-based brain age frameworks: A
  robust scheme,''
\newblock {\em NeuroImage: Clinical}, vol. 24, pp. 102063, 2019.

\bibitem{desikan2009temporoparietal}
Rahul~S Desikan, Howard~J Cabral, Bruce Fischl, Charles~RG Guttmann, Deborah
  Blacker, Bradley~T Hyman, Marilyn~S Albert, and Ronald~J Killiany,
\newblock ``Temporoparietal mr imaging measures of atrophy in subjects with
  mild cognitive impairment that predict subsequent diagnosis of alzheimer
  disease,''
\newblock {\em American Journal of Neuroradiology}, vol. 30, no. 3, pp.
  532--538, 2009.

\bibitem{rohrer2012structural}
Jonathan~D Rohrer,
\newblock ``Structural brain imaging in frontotemporal dementia,''
\newblock {\em Biochimica et Biophysica Acta (BBA)-Molecular Basis of Disease},
  vol. 1822, no. 3, pp. 325--332, 2012.

\bibitem{mcfarland2016diagnostic}
Nikolaus~R McFarland,
\newblock ``Diagnostic approach to atypical parkinsonian syndromes,''
\newblock {\em CONTINUUM: Lifelong Learning in Neurology}, vol. 22, no. 4, pp.
  1117--1142, 2016.

\bibitem{lombardi2021explainable}
Angela Lombardi, Domenico Diacono, Nicola Amoroso, Alfonso Monaco, Jo{\~a}o
  Manuel~RS Tavares, Roberto Bellotti, and Sabina Tangaro,
\newblock ``Explainable deep learning for personalized age prediction with
  brain morphology,''
\newblock {\em Frontiers in neuroscience}, vol. 15, pp. 578, 2021.

\bibitem{yin2023anatomically}
Chenzhong Yin, Phoebe Imms, Mingxi Cheng, Anar Amgalan, Nahian~F Chowdhury,
  Roy~J Massett, Nikhil~N Chaudhari, Xinghe Chen, Paul~M Thompson, Paul Bogdan,
  et~al.,
\newblock ``Anatomically interpretable deep learning of brain age captures
  domain-specific cognitive impairment,''
\newblock {\em Proceedings of the National Academy of Sciences}, vol. 120, no.
  2, pp. e2214634120, 2023.

\bibitem{hofmann2022towards}
Simon~M Hofmann, Frauke Beyer, Sebastian Lapuschkin, Ole Goltermann, Markus
  Loeffler, Klaus-Robert M{\"u}ller, Arno Villringer, Wojciech Samek, and
  A~Veronica Witte,
\newblock ``Towards the interpretability of deep learning models for
  multi-modal neuroimaging: Finding structural changes of the ageing brain,''
\newblock {\em NeuroImage}, vol. 261, pp. 119504, 2022.

\end{thebibliography}
}
\end{document}